\documentclass[10pt,conference,a4paper]{IEEEtran}
\IEEEoverridecommandlockouts
\usepackage{cite}
\usepackage{amsmath,amssymb,amsfonts}
\usepackage{algorithmic}
\usepackage{graphicx}

\def\BibTeX{{\rm B\kern-.05em{\sc i\kern-.025em b}\kern-.08em
    T\kern-.1667em\lower.7ex\hbox{E}\kern-.125emX}}
\ifCLASSOPTIONcompsoc
  \usepackage[caption=false,font=normalsize,labelfont=sf,textfont=sf]{subfig}
\else
  \usepackage[caption=false,font=footnotesize]{subfig}
\fi
\begin{document}

\title{Modulation Pattern Detection Using Complex Convolutions in Deep Learning}
\author{\IEEEauthorblockN{Jakob Krzyston}
\IEEEauthorblockA{\textit{School of Electrical and Computer Engineering} \\
\textit{Georgia Institute of Technology}\\
Atlanta, USA \\
jakobk@gatech.edu}
\and
\IEEEauthorblockN{Rajib Bhattacharjea}
\IEEEauthorblockA{\textit{DeepSig Inc.} \\
Arlington, USA\\
rbhattacharjea@deepsig.io
}
\and
\IEEEauthorblockN{Andrew Stark}
\IEEEauthorblockA{\textit{Electro-Optical Systems Laboratory} \\
\textit{Georgia Tech Research Institute}\\
Atlanta, USA\\
andy.stark@gtri.gatech.edu
}
}

\maketitle

\begin{abstract}
Transceivers used for telecommunications transmit and receive specific modulation patterns that are represented as sequences of complex numbers. Classifying modulation patterns is challenging because noise and channel impairments affect the signals in complicated ways such that the received signal bears little resemblance to the transmitted signal. Although deep learning approaches have shown great promise over statistical methods in this problem space, deep learning frameworks continue to lag in support for complex-valued data. To address this gap, we study the implementation and use of complex convolutions in a series of convolutional neural network architectures. Replacement of data structure and convolution operations by their complex generalization in an architecture improves performance, with statistical significance, at recognizing modulation patterns in complex-valued signals with high SNR after being trained on low SNR signals. This suggests complex-valued convolutions enables networks to learn more meaningful representations. We investigate this hypothesis by comparing the features learned in each experiment by visualizing the inputs that results in one-hot modulation pattern classification for each network.

\end{abstract}

\section{Introduction}
Transceivers used for telecommunications transmit and receive sequences of data that are conventionally thought of as representing complex numbers. In the field of radio-frequency signals, the real and imaginary components of these complex data are called the in-phase (I) and quadrature (Q) components respectively. Predetermined modulation patterns provide unique structure that can be decoded by the receiver to reproduce the intended message, more details on this can be found in Section II. Despite the intentional structure and pattern in I/Q signals, there are sources of noise and distortion in wireless channels, leading to errors in decoding the data, and making simple modulation classification particularly challenging. For many years statistical machine learning approaches were used for I/Q modulation classification both in research and in the field, but these methods struggle as SNR decreases~\cite{isautier2015agnostic,isautier2015stokes}.

In recent years, the machine learning field has shifted to deep learning approaches due to remarkable performance on machine learning tasks such as image classification~\cite{NIPS2012_4824,huang2018gpipe} and playing board games~\cite{silver2017mastering}. Of these approaches, convolutional neural networks (CNNs) have had the greatest impact on the telecommunications field for modulation classification~\cite{o2016convolutional}. However, the traditional implementation of a CNN convolution does not properly compute a complex convolution. In previous work~\cite{krzyston2020complex}, we demonstrated a novel method for performing convolutions on complex-valued data and showed this technique performs up to 35\% better than traditional CNN approaches at detecting modulation pattern structure in the presence of noise. 

In this paper, we expand on our previous work by utilizing complex convolution-based CNNs to detect patterns in I/Q modulated signals in various training and testing scenarios, varying the amount of noise in the training and testing sets. In addition to reporting modulation pattern classification accuracy, we analyze the features learned in each train/test scenario to better understand what patterns the networks have learned. To do this we compute the input, for a given trained network, that results in 100\% classification accuracy for that modulation class, and simultaneously 0\% classification accuracy for all other modulation classes, termed ‘one-hot’ classification.

\begin{figure}
 \centering
  \subfloat[\label{fig:IQ_1}]{%
       \includegraphics[width = 1.9 in]{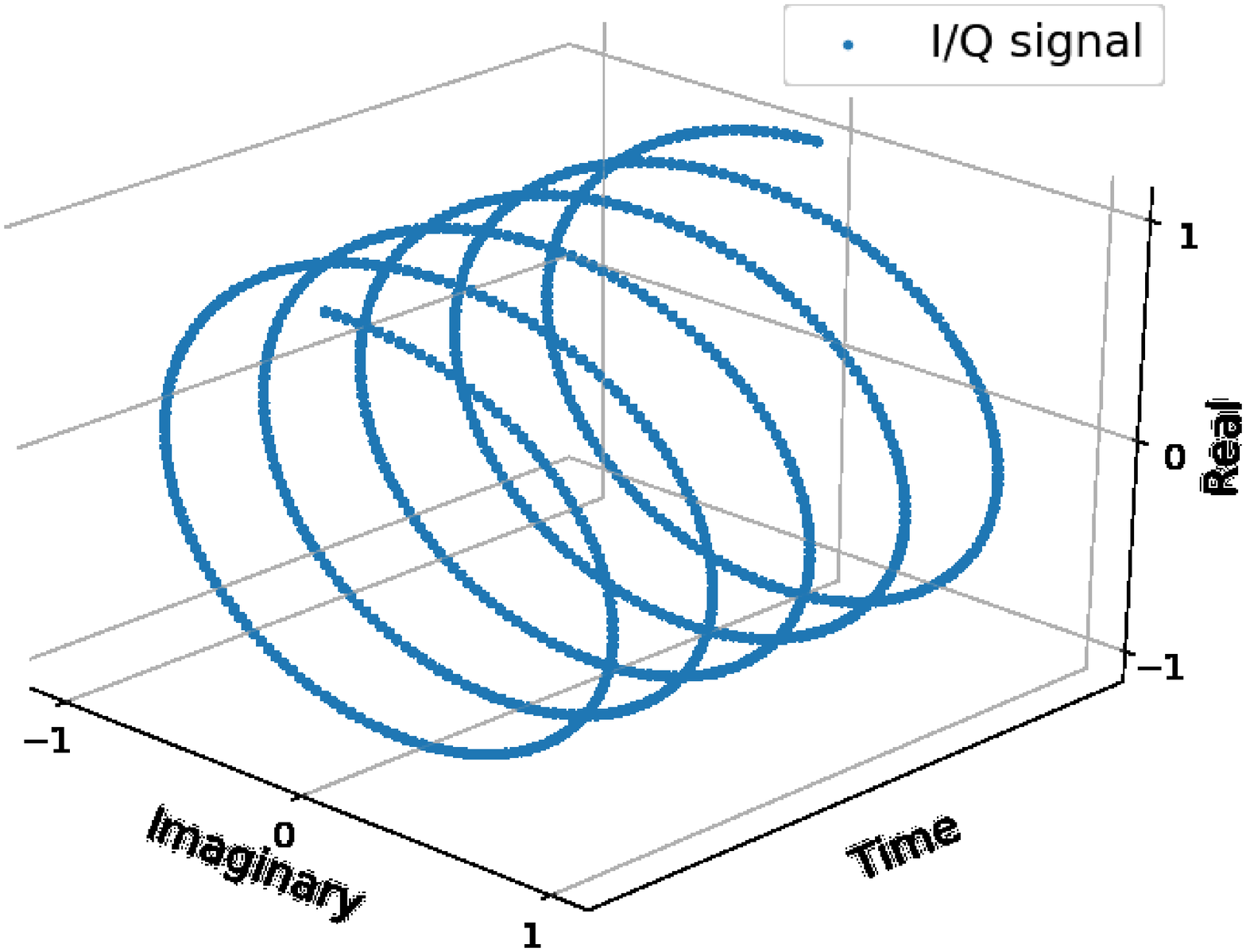}}
  \hfill
  \subfloat[\label{fig:IQ_2}]{%
        \includegraphics[width = 1.5 in]{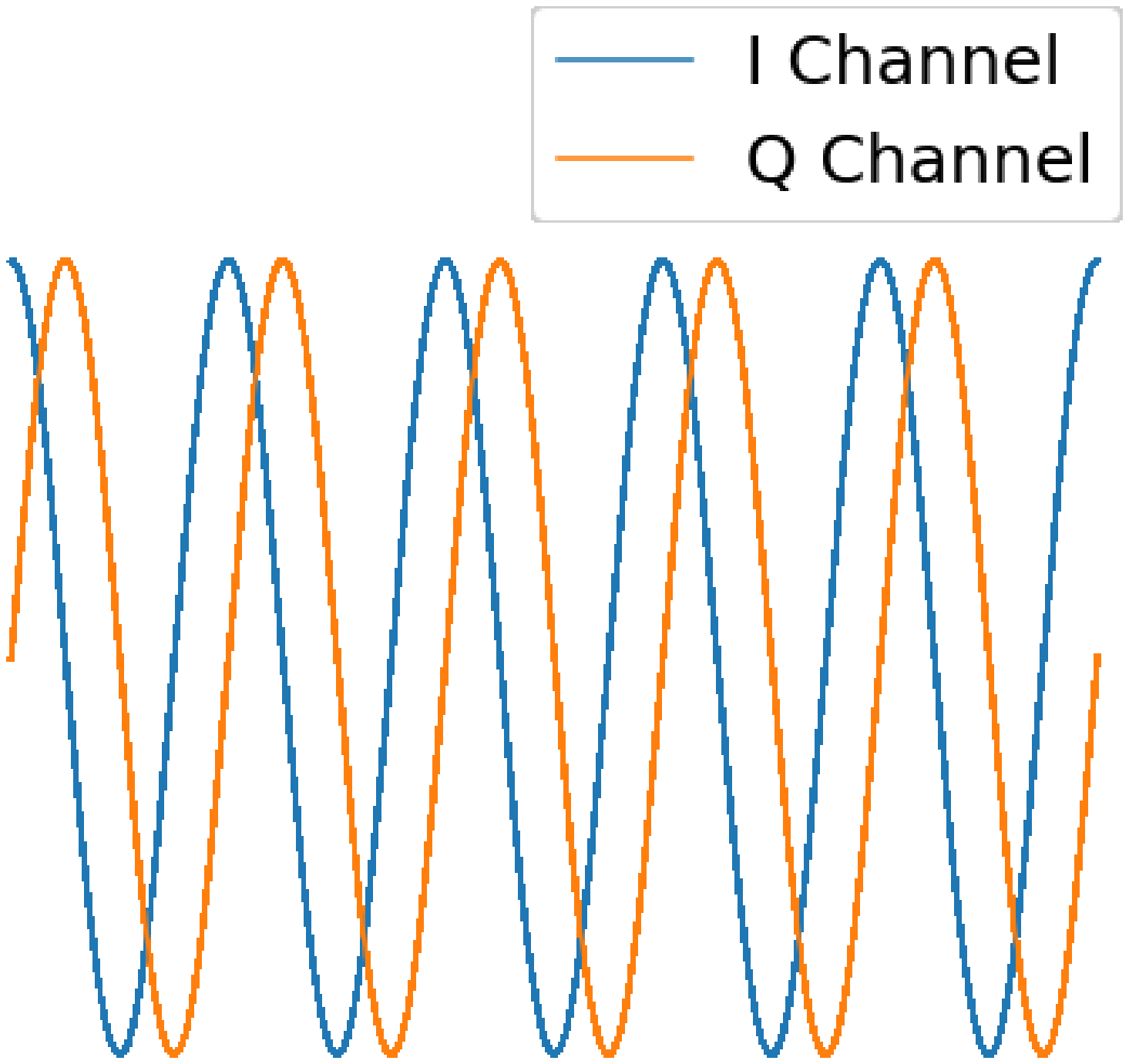}}
\caption{(a) Example I/Q signal, $e^{j\phi(t)}$, shown in three dimensional space. (b) A two dimensional view of the example I/Q signal shown in \ref{fig:IQ_1}. The blue plot shows the I channel over time, $\cos(\phi(t))$, and the orange plot shows the Q channel over time, $\sin(\phi(t))$. In this report, this visualization will be used to display I and Q channels of a given I/Q sample.}
\end{figure}

\section{I/Q Modulation Overview}
I/Q signals can be represented in Euler form as $A(t)e^{j\phi(t)}$, where $A(t)$ is the real-valued time-dependent magnitude, $j$ the imaginary unit ($j=\sqrt{-1}$), and $\phi(t)$ the real-valued angle of rotation as a function of time, Figure~\ref{fig:IQ_1}. This signal can be broken down into components using Euler's identity, $A(t)e^{j\phi(t)} = A(t)\cos(\phi(t))+jA(t)\sin(\phi(t))$, where the I component is $A(t)\cos(\phi(t))$ and the Q component is $A(t)\sin(\phi(t))$ as shown in Figure~\ref{fig:IQ_2}.

\section{Deep Learning for Modulation Classification}
O'Shea et al. \cite{o2016convolutional} first introduced deep learning approaches to I/Q modulation pattern classification, showing great accuracy improvements over traditional statistical methods. In \cite{o2018over}, a thorough exercise in various methods for radio signal classification was conducted. The results show deep learning architectures such as ResNets significantly outperform advanced statistical machine learning methods such as gradient boosted trees with hand crafted high-order statistical features. In \cite{ramjee2019fast}, Ramjee et al. furthered this work by trying many different deep learning approaches to classify modulated radio signals and investigated as how to best reduce the training time of the approaches. Deep learning architectures investigated include their own CNN, DenseNet, and Convolutional Long Short-term Deep Neural Network (CLDNN) architectures, as well as hyperparameter tuned versions of the ResNet and the LSTM architectures. Their results indicated tuned versions of the ResNet and LSTM performed better at different SNRs.

Ramjee et al. demonstrated that, in general, more sophisticated architectures outperform those demonstrated by O'Shea et al. since they are capable of mapping new non-linear relationships in addition to having many more parameters to optimize over.

This paper will leverage the simplicity of more traditional CNN architectures, as seen in O'Shea et al., to emphasize the benefits of using our complex convolution approach. Future work will investigate sophisticated architectures, complex-valued activation functions, and batch-normalizations.

\section{Current Methods for Handling Complex Numbers in Deep Learning}
In deep learning, CNNs in particular, there are a few common approaches to learn features on multi-channeled data (e.g. complex-valued data, RGB images)
\begin{itemize}
\item[1.] Combine all channels via a 1 x 1 convolution and learn features over the reduced-dimension image \cite{GoogLeNet}
\item[2.] Learn a set of features per channel
\item[3.] Learn one set of features for all channels (i.e. both the I and Q components as seen in \cite{o2016convolutional}) 
\end{itemize} 
However, these approaches do not account for patterns that arise from the modulation-specific relationship between the I and Q channels.

As previously detailed in \cite{krzyston2020complex}, there have been other works which have allowed neural networks to perform computations with complex numbers~\cite{arjovsky2016unitary,virtue2017better,trabelsi2017deep} but do not directly compute complex convolutions on complex-valued inputs.

Recently, there has been another approach developed to enable deep learning paradigms to compute complex convolutions~\cite{chakraborty2019surreal}. 
This work takes a geometric approach to understanding the relationship between the real and imaginary components by defining the convolution as a weighted Fr\'{e}chet mean on a Lie group. This new form of convolution necessitated the development of a new activation function, $G$-transport.

\section{Complex Convolution Using 2D Real Convolution}

In this section we detail how to compute a complex convolution in real-valued deep learning frameworks as shown in \cite{krzyston2020complex}. 

Consider a two dimensional I/Q signal $(Z_n)_{n=1}^N$.
The elements of $Z_n$ are defined by
$$
Z_n = I_n + jQ_n, \; I_n,Q_n\in\mathbb{R}
$$
where $I_n$ and $Q_n$ represent the $n^{th}$ in-phase and quadrature components of $Z$. $I$ and $Q$ contain all $N$ elements in the respective channels and can be arranged in an $N$x2 array, as shown.

\begin{equation}\label{eq:IQData}
\begin{array}{|c|c|}
\hline
I_1 & Q_1\\\hline
I_2 & Q_2\\\hline
I_3 & Q_3\\\hline
\vdots & \vdots \\\hline
I_N & Q_N\\\hline
\end{array}
\end{equation}

We introduce another I/Q signal, $h$, that contains $M$ complex filter coefficients,
$$
h_m = h_m' + jh_m''\;\;
$$
where $h_m'$ and $h_m''$ are the $m^{th}$ in-phase and quadrature components of $h$. The sequence of coefficients $h$ can be thought of as weights in a convolutional filter.

In deep learning, convolutions act as a na\"ive sliding window performing cross-correlation, thus the 'convolution' of $Z$ and $h$ yields $X_\text{DL}$, which contains three columns as shown in Equations \ref{eq:realConv2D} \& \ref{eq:realConv2D_pt2}.

\begin{equation}\label{eq:realConv2D}
X_\text{DL} = 
\begin{array}{|c|c|}
\hline
I_1 & Q_1\\\hline
I_2 & Q_2\\\hline
I_3 & Q_3\\\hline
\vdots & \vdots \\\hline
I_N & Q_N\\\hline
\end{array}
*
\begin{array}{|c|c|}
\hline
h_1' & h_1''\\\hline
h_2' & h_2''\\\hline
h_3' & h_3''\\\hline
\vdots & \vdots \\\hline
h_M' & h_M''\\\hline
\end{array}
\end{equation}

\begin{equation}\label{eq:realConv2D_pt2}
X_\text{DL}
=
\begin{array}{|c|c|c|}
\hline
\uparrow & \uparrow & \uparrow  \\
I*h' & I*h''+ Q*h' & Q*h'' \\
\downarrow & \downarrow & \downarrow \\
\hline
\end{array}
\end{equation}

Where $*$ denotes convolution.

By means of linear combination, we can  reconstruct the two channel array with the proper complex convolution calculated by subtracting the third column from the first, shown in Equation \ref{eq:complexConv1D}.
\begin{equation}\label{eq:complexConv1D}
X = 
\begin{array}{|c|c|}
\hline
\uparrow & \uparrow \\
I*h'-Q*h'' & I*h'' + Q*h' \\
\downarrow & \downarrow \\
\hline
\end{array}
\end{equation}

\begin{figure*}[!htb]
 \centering
  \subfloat[\label{fig:CNN2}]{%
       \includegraphics[width=2.9in]{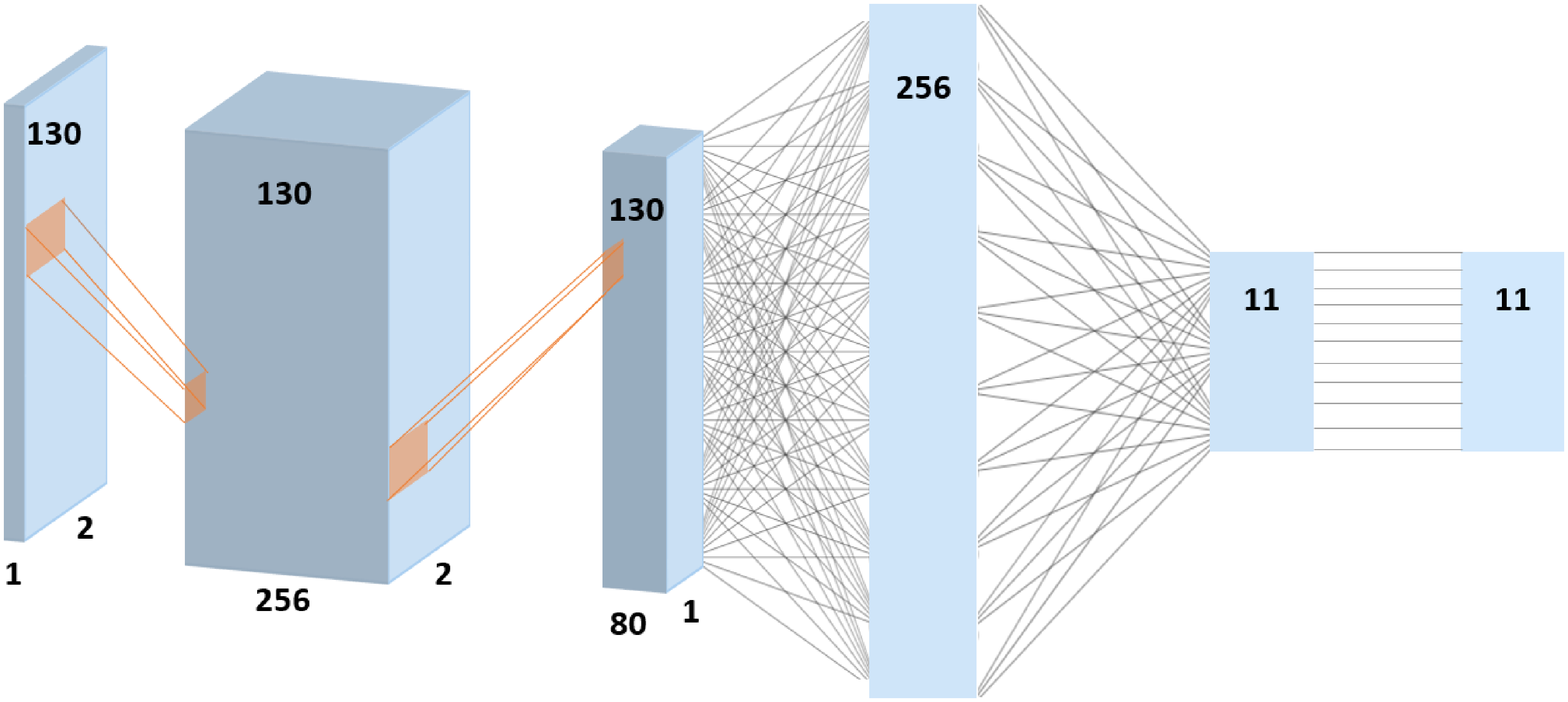}}
    \hfill
  \subfloat[\label{fig:complex}]{%
        \includegraphics[width=4.1in]{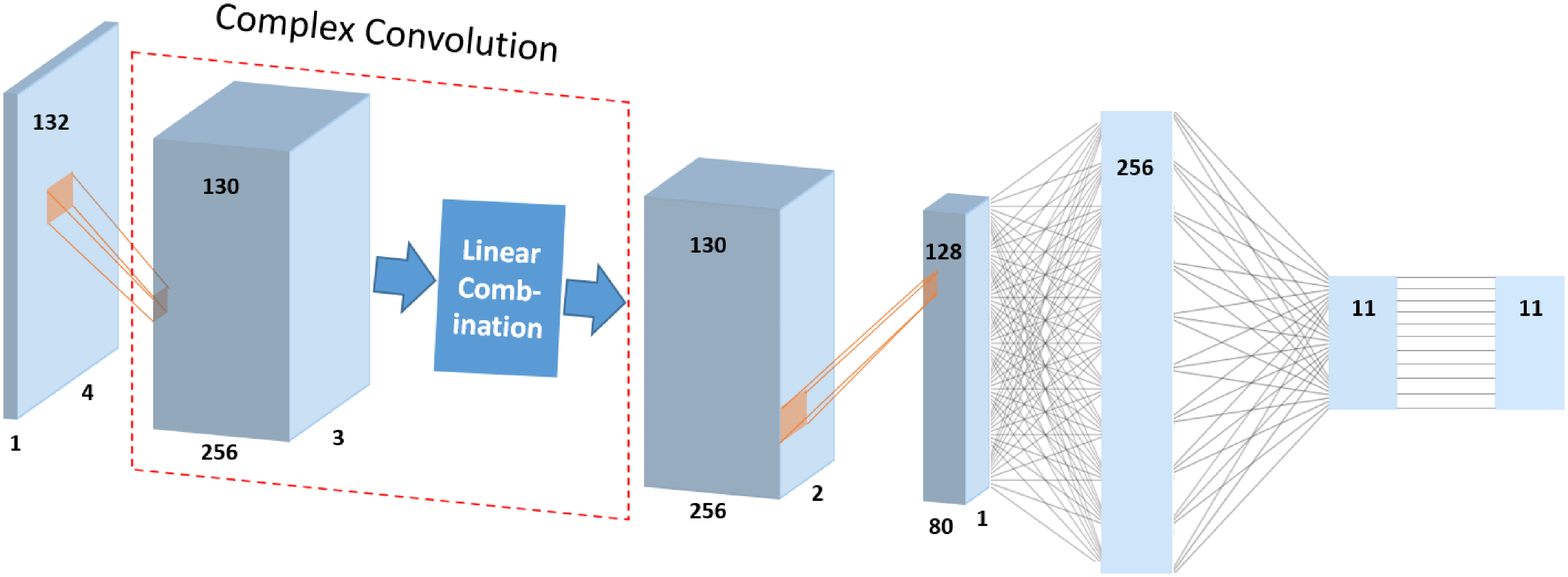}}
\caption{(a) Schematic of the CNN2 architecture used to classify I/Q modulation patterns in \cite{o2016convolutional}. (b) Schematic of the Complex architecture. For further details of both architectures, please refer to \cite{krzyston2020complex}.}
\end{figure*}

\section{Deep Learning Architectures}
To better demonstrate the abilities of the complex convolution, we use the three simple CNN architectures used in \cite{krzyston2020complex}.
\subsection{CNN2}
The baseline CNN used in this study was first used in \cite{o2016convolutional}, and was named `CNN2'. It is comprised of two convolutional layers, two dense layers, and a multiclass softmax layer. Shown in Figure~\ref{fig:CNN2}, CNN2 is a CNN that has been used to classify modulated I/Q signals. 

\subsection{Complex}
In \cite{krzyston2020complex} a linear combination was implemented to compute a complex valued convolution. This is implemented into CNN2 to become the architecture we named `Complex', seen in  Figure~\ref{fig:complex}. The linear combination increases the ability for pattern recognition as it allows the entire I/Q signal two-channel structure to be passed along the network. The Complex architecture we present allows for increased filter size that improves the ability of the network to identify unique patterns in and between the I and Q channels, which is not done in \cite{o2016convolutional}.

\subsection{CNN2-257}
To ensure a fair comparison between the Complex and CNN2 architectures, we added parameters to CNN2 to compensate for the number of parameters in the Complex network. The modification was to increase the dense layer size from 256 to 257, hence CNN2-257. The CNN-257 network has 0.3\% more parameters than the Complex architecture, which we deem appropriate to demonstrate the difference between simply adding more parameters to increase performance and being able to compute a complex convolution.

All architectures used a dropout value of 0.5, ReLU activation function, Adam optimizer, and categorical cross-entropy loss function. Further details for all networks can be found in \cite{krzyston2020complex}.

\begin{figure}
\centering
\includegraphics[width=3.45 in]{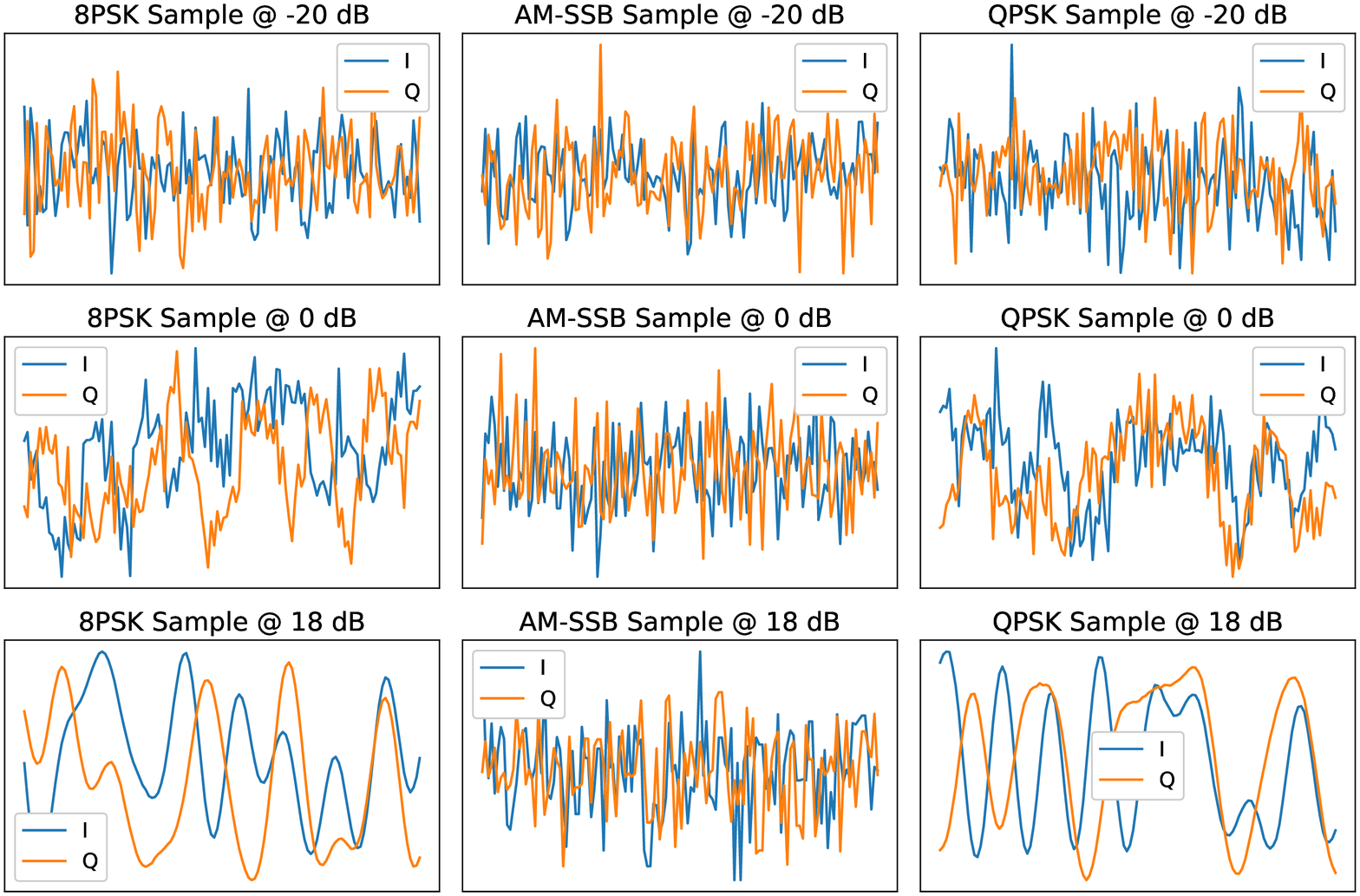}
\caption{I/Q signal samples from the dataset. Top to bottom is the variation in SNR (-20, 0, and 18 db), while left to right varies the modulation pattern (8PSK, AM-SSB, and QPSK). Each panel shows 128 samples of the I and Q channels.}
\label{fig:examples}
\end{figure}

\section{IQ Modulation Classification Task}
These architectures were trained and tested on the RadioML 2016.10A open source dataset used in \cite{o2016convolutional}. The dataset  consists of 11 modulations (8 digital and 3 analog) at SNR levels from -20 to 18 dB with 2 dB steps. Additionally the dataset includes variation in the following properties: center frequency, sample clock rate, sample clock offset, and initial phase. There are 1,000 examples of all modulation schemes at all SNR values in the entire dataset. The shape of each example is 2 x 128 \cite{o2016convolutional}.

Examples of a few modulations such as Eight Phase Shift Keying (8PSK), Amplitude-Modulated Single-Sideband Modulation (AM-SSB), Binary Phase Shift Keying (BPSK), and Quadrature Amplitude Modulation 16 (QAM16) are shown in Figure~\ref{fig:examples} at -20, 0, and 18 dB SNR. These examples highlight the effects of these noise models as a function of SNR, as well as display the stark difference that exists in the ability to readily see structure in each modulation pattern.

\begin{figure*}
 \centering
  \subfloat[\label{fig:class_me}]{%
       \includegraphics[width = 2.35 in]{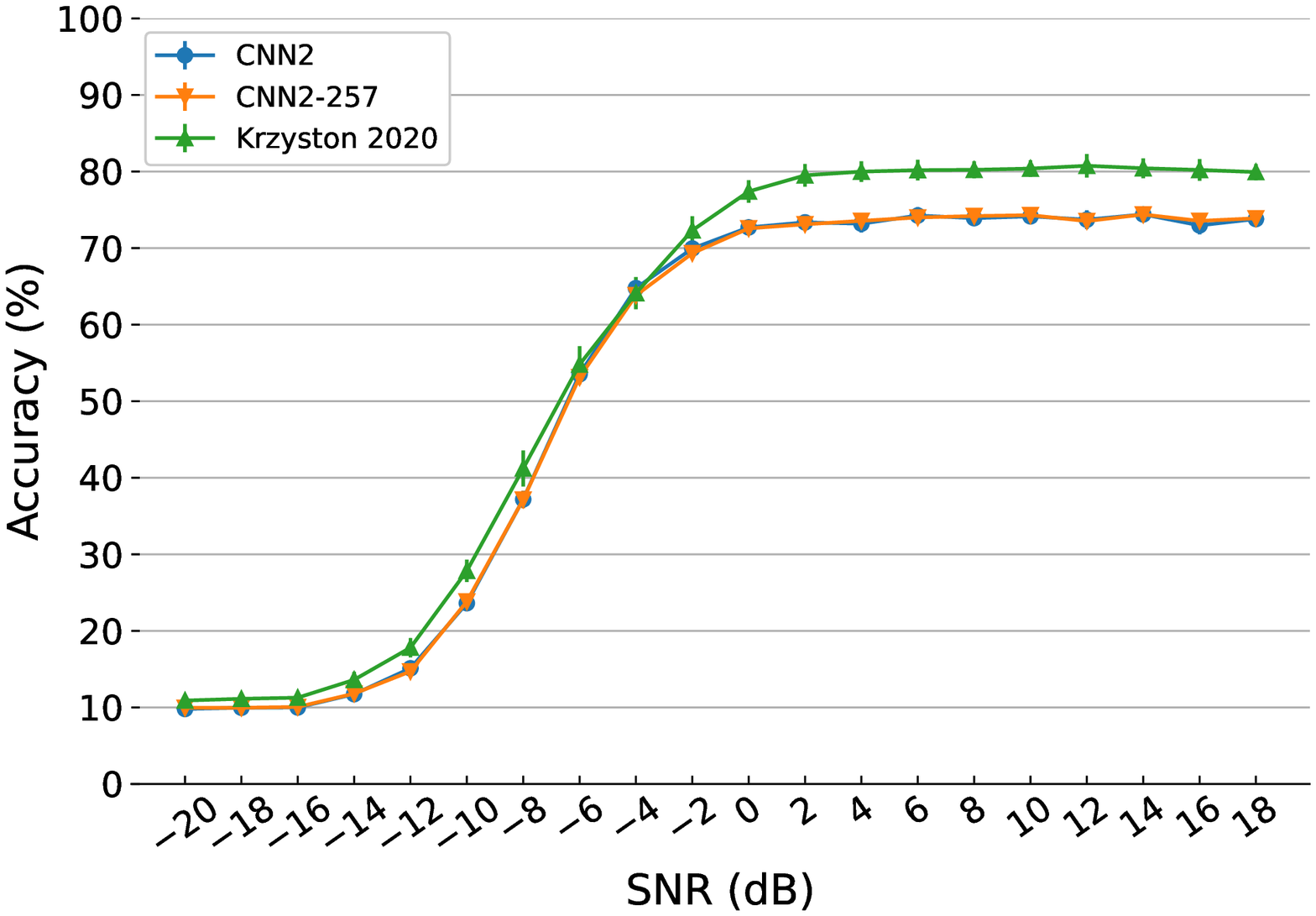}}
  \hfill
  \subfloat[\label{fig:class_1}]{%
        \includegraphics[width = 2.35 in]{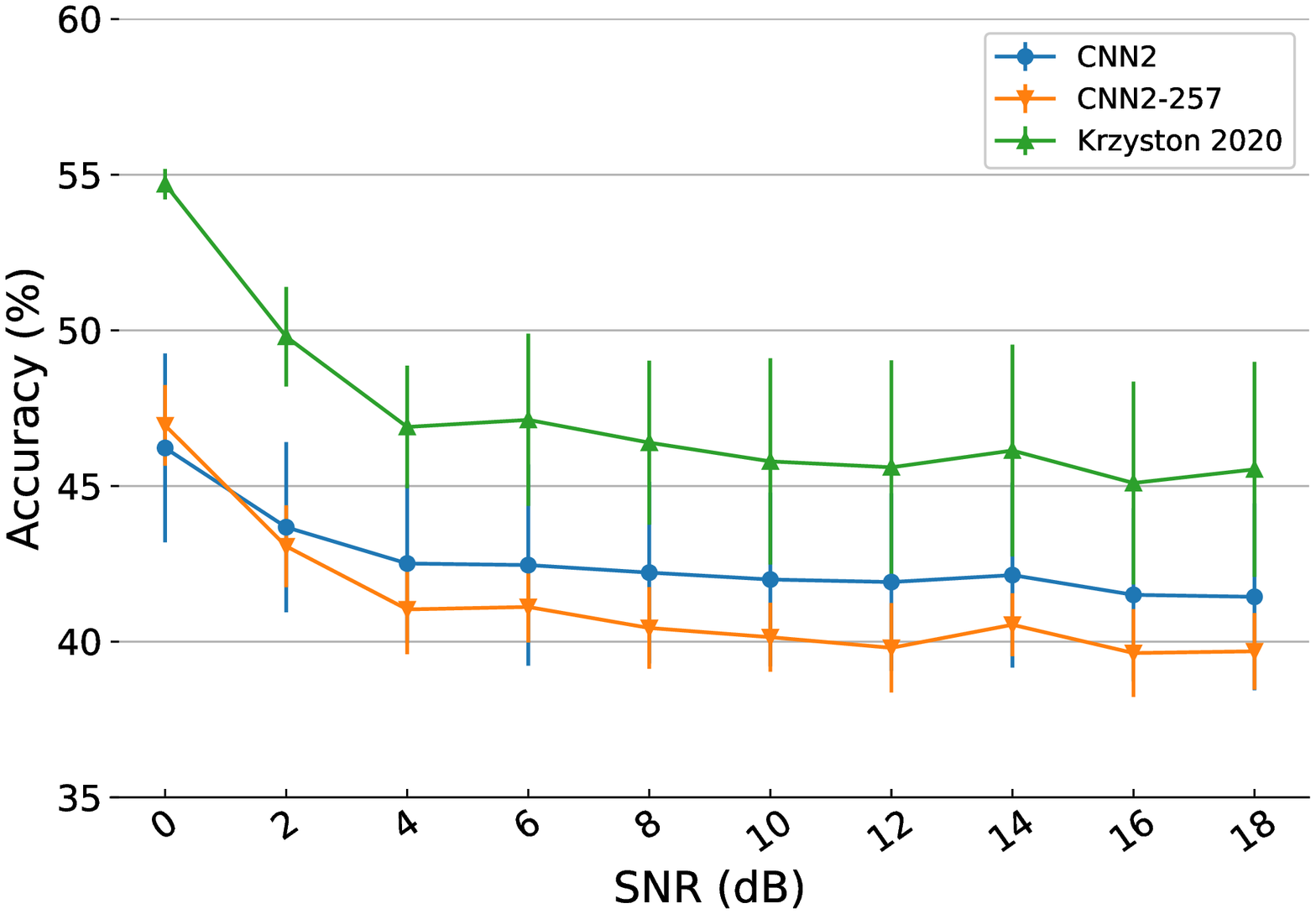}}
  \hfill  
  \subfloat[\label{fig:class_2}]{%
       \includegraphics[width = 2.35 in]{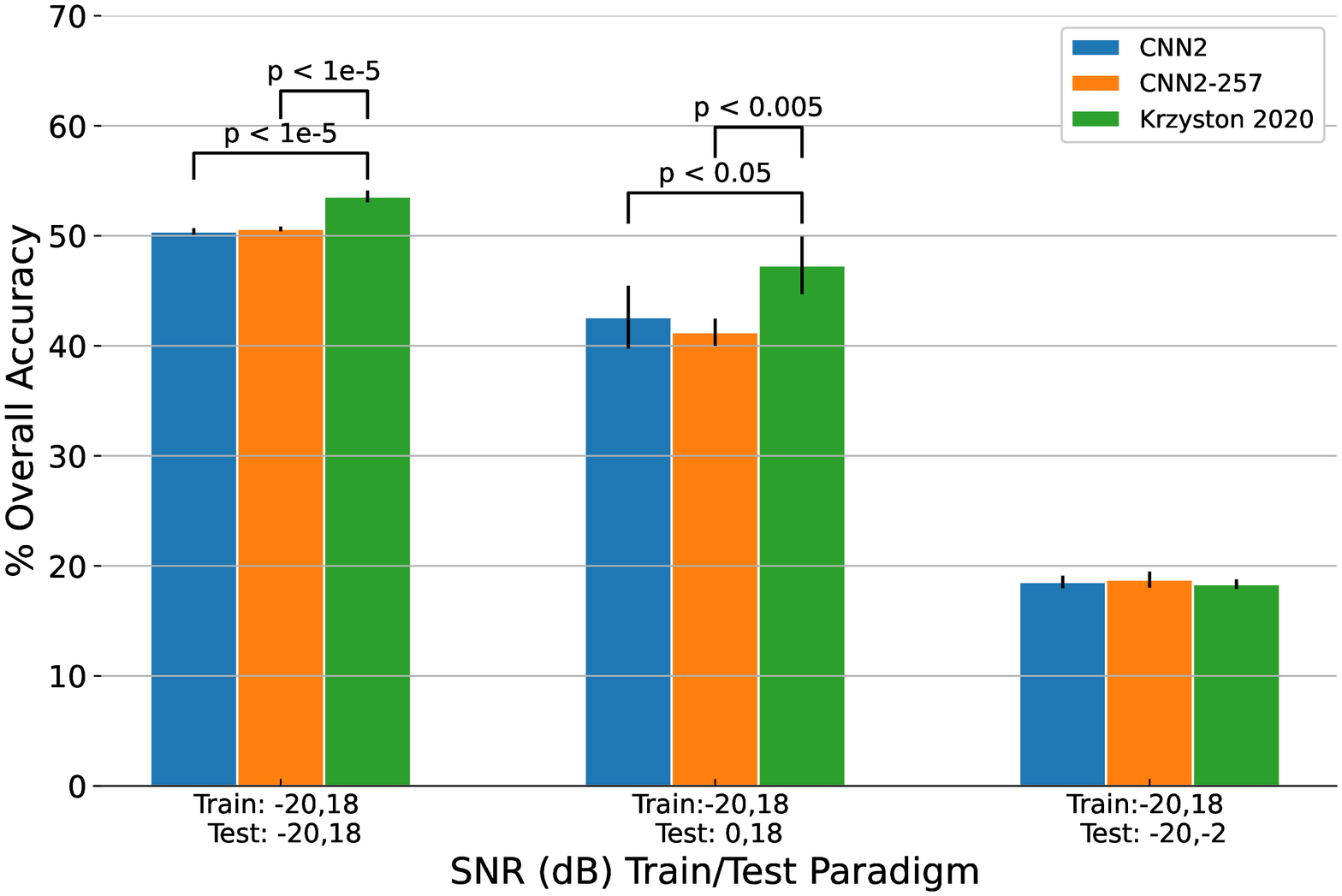}}
\caption{Classification accuracy plots as a function of SNR, with the addition of standard deviation bars over the five trials, of (a) the experiment performed in \cite{krzyston2020complex}, and (b) Experiment 1 the networks were trained on [-20, -2] db SNR data and tested on [0, 20] db SNR data. Figure 4(c) displays the average accuracy, over all modulations and SNRs, and standard deviation for each experiment, along with p-values comparing the performances amongst the architectures, if statistically significant. The unpaired student t-test was used to compute the p-values.}
\label{fig:accs}
\end{figure*}

\section{Experimental Design}
In \cite{krzyston2020complex}, these architectures were trained and tested on all SNRs of all modulations, with a 50/50 test/train split each time. In this work, two new experiments were proposed to demonstrate the ability of complex convolutions to learn pattern-based features in the presence of noise: 
\begin{itemize}
\item[1.] Train: [-20, -2] dB SNR data, Test: [0, 18] dB SNR data
\item[2.] Train: [0, 18] dB SNR data, Test: [-20, -2] dB SNR data
\end{itemize}
In accordance with the original experiment \cite{krzyston2020complex}, the training and testing data were shuffled across modulation type and SNR, and contained the same number of samples, 110,000. 
Unlike \cite{krzyston2020complex}, we repeat the experiments five times in order to determine statistical significance in performances, this necessitated repeated experiments of what was published in \cite{krzyston2020complex}.

\section{Results}
\subsection{Classification Results}
In Figures~\ref{fig:accs}b and c we see the Complex network outperformed the CNN2 and CNN2-257 architectures with statistical significance in the same experiment performed in~\cite{krzyston2020complex} and  Experiment 1, while the results were statistically equivalent for Experiment 2. The addition of the linear combination increased accuracy at all SNR levels in Experiment 1, in accordance with the results from \cite{krzyston2020complex}. Figure~\ref{fig:accs}a shows modulation pattern classification with the Complex network exceed 80\% at SNRs above 2 dB. In Figure~\ref{fig:accs}b the Complex network averages over 45\% classification accuracy for all tested SNR values whereas the other networks average 45\% at only one tested SNR. Figure~\ref{fig:accs}c shows the average classification accuracy for each architecture for each experiment, along with standard deviation bars. Further, Figure~\ref{fig:accs}c is annotated with p-values showing statistical improvement in classification performance from the Complex network over the other architectures in the train/test paradigms from \cite{krzyston2020complex} and Experiment 1. The results from Experiment 2 were deemed to be statistically inconclusive, as all architectures performed poorly when classifying the low SNR data after being trained on the higher SNR data.

\begin{figure*}
 \centering
  \subfloat[\label{fig:conf_me}]{%
       \includegraphics[width = 2.3 in]{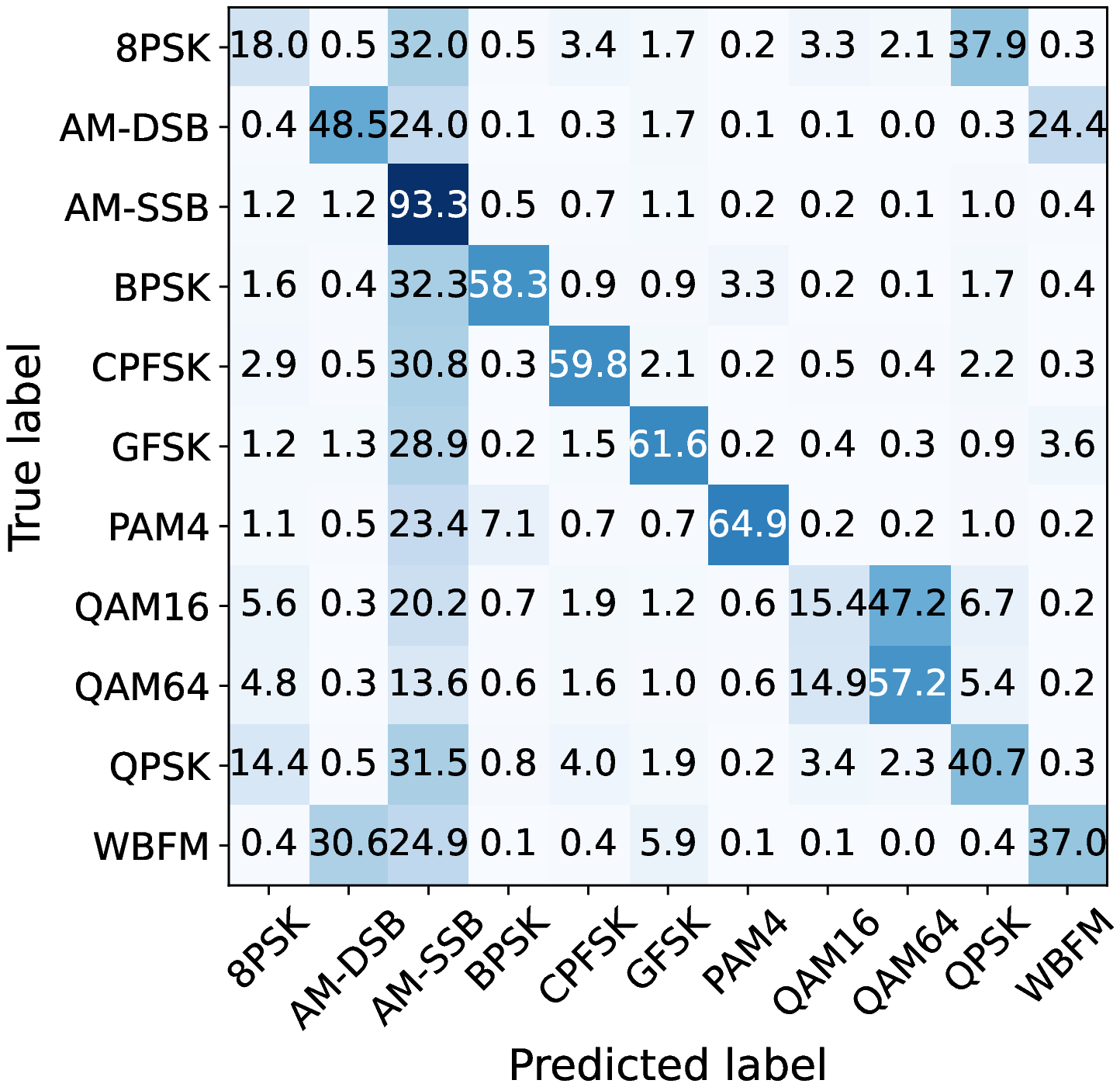}}
  \hfill
  \subfloat[\label{fig:conf_1}]{%
        \includegraphics[width = 2.3 in]{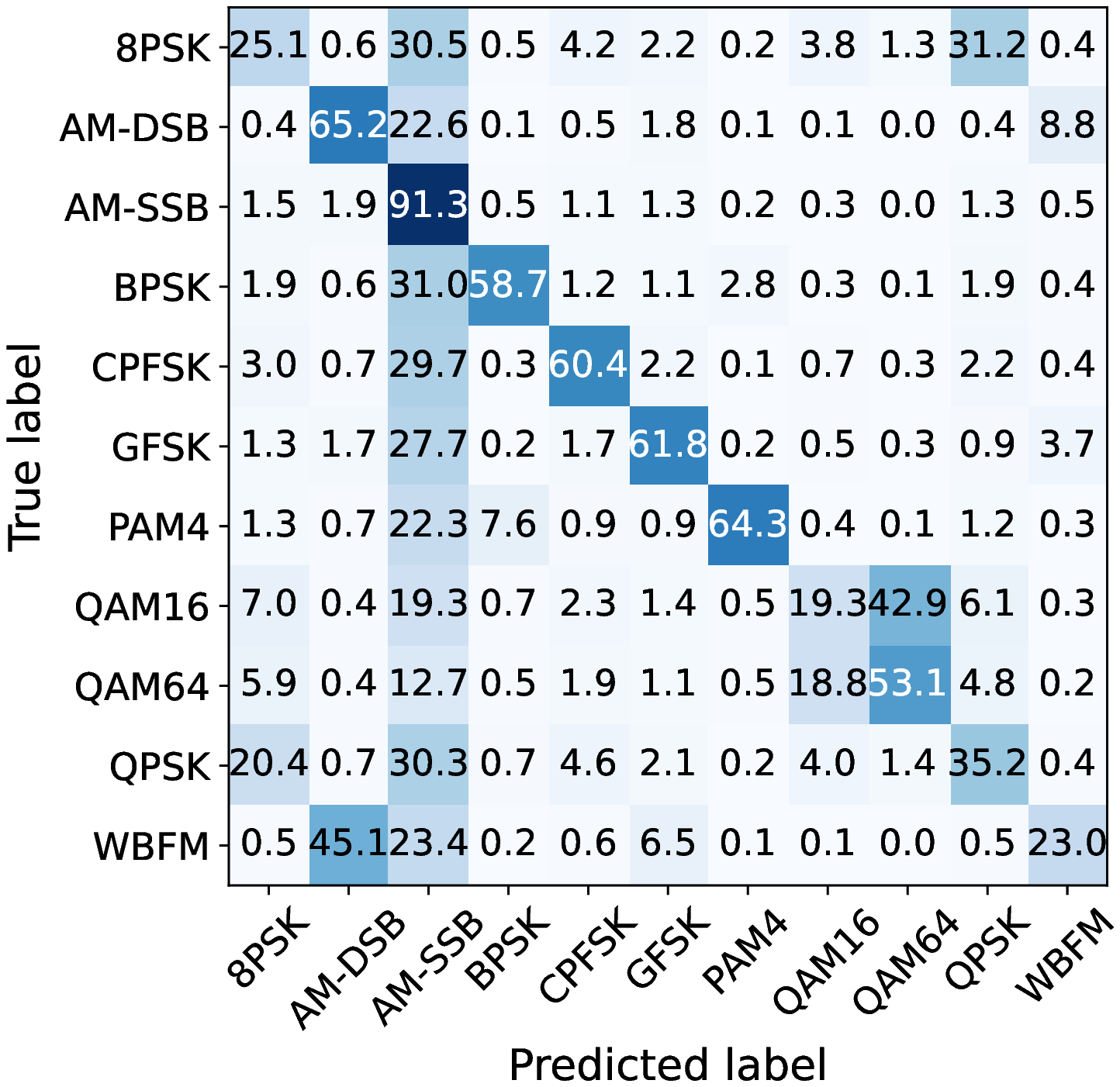}}
  \hfill  
  \subfloat[\label{fig:conf_2}]{%
       \includegraphics[width = 2.3 in]{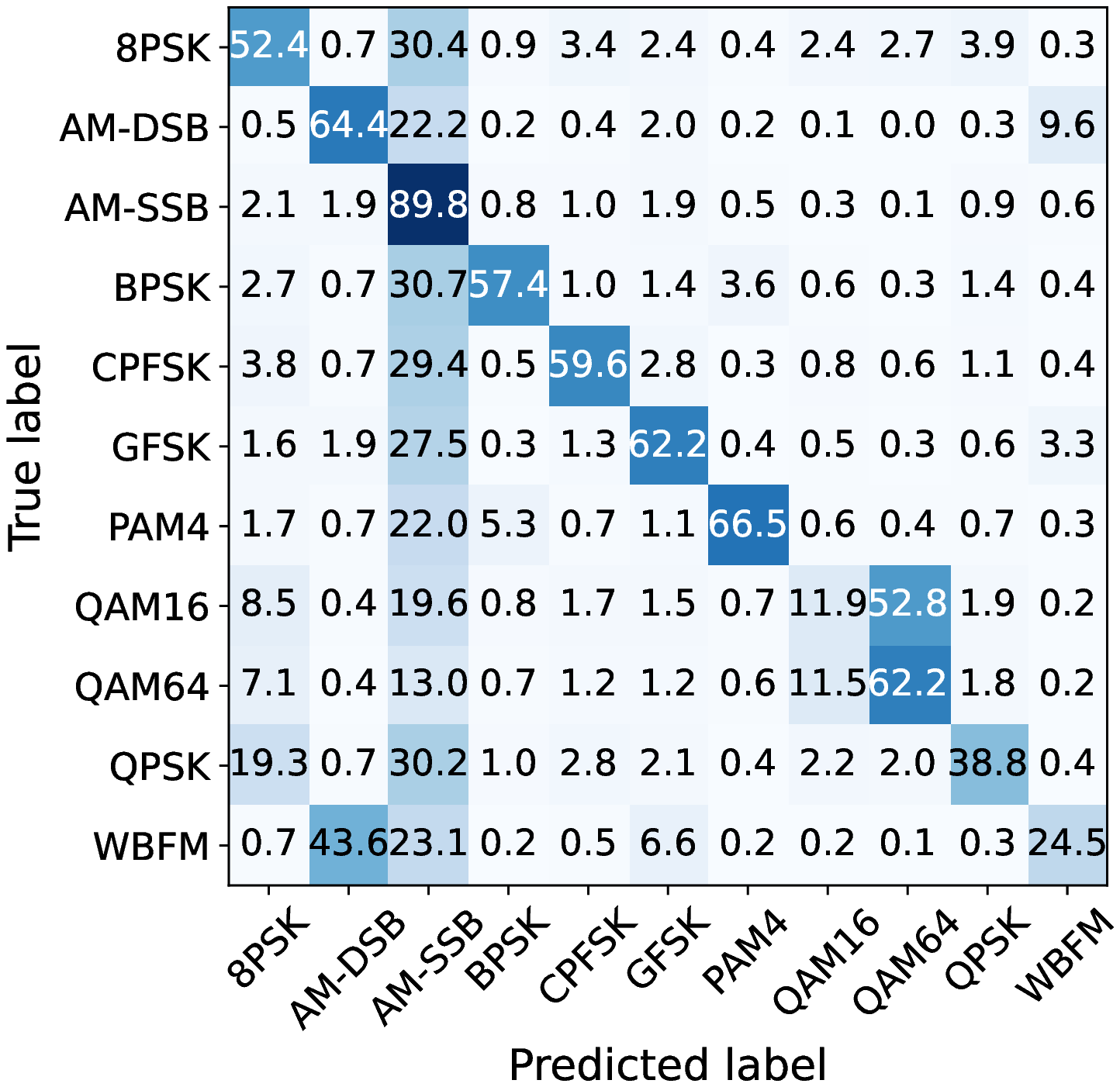}}
\caption{Overall average confusion matrices of the experiment run in \cite{krzyston2020complex} showing the performance of the (left) CNN2, (middle) CNN2-257, and (right) Complex architectures. The respective average modulation pattern classification accuracies are 50.39\%, 50.62\%, and 53.57\%.}
\label{fig:conf}
\end{figure*}

\begin{figure*}
 \centering
  \subfloat[\label{fig:conf1_me}]{%
       \includegraphics[width = 2.3 in]{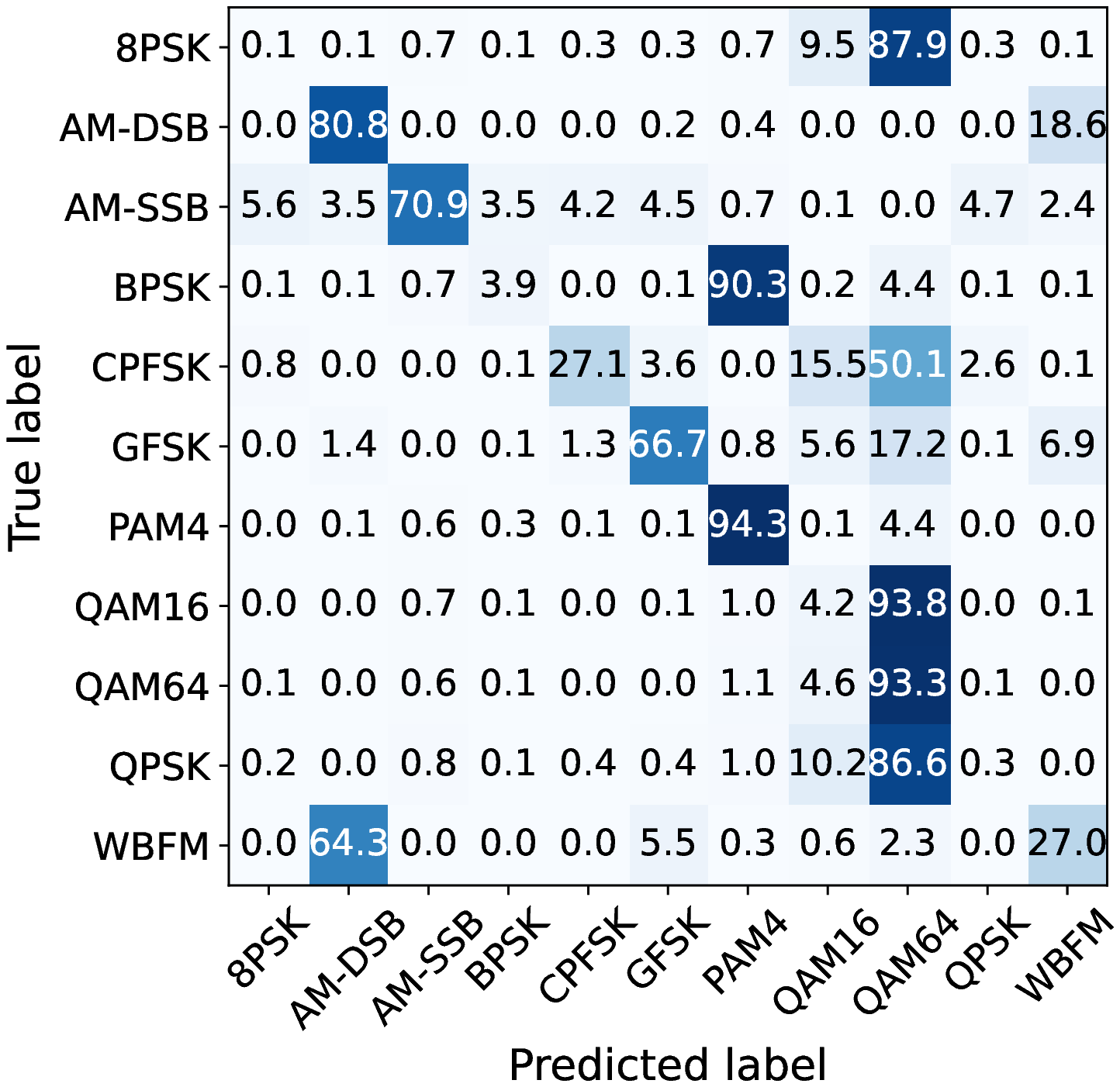}}
  \hfill
  \subfloat[\label{fig:conf1_1}]{%
        \includegraphics[width = 2.3 in]{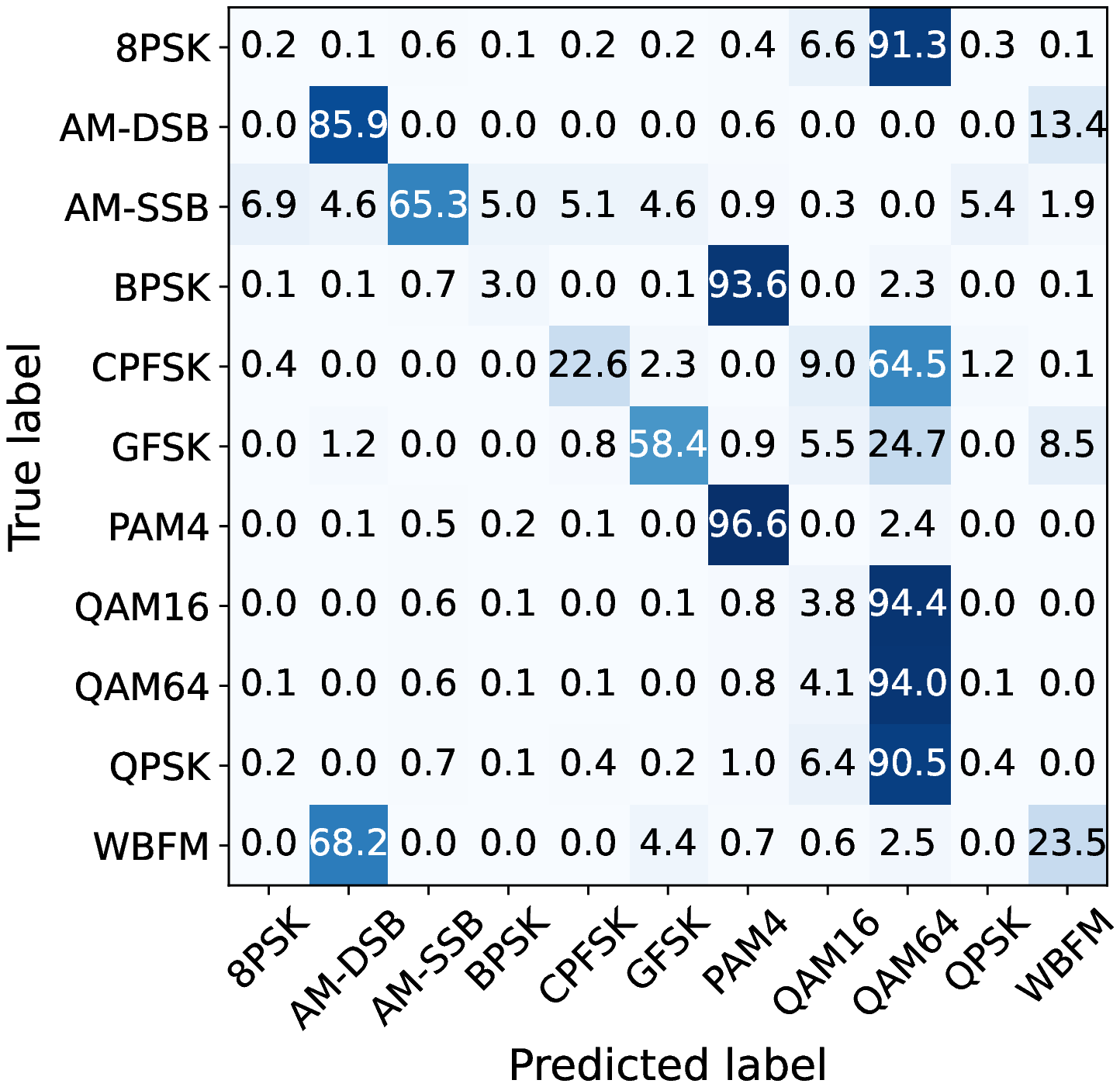}}
  \hfill  
  \subfloat[\label{fig:conf1_2}]{%
       \includegraphics[width = 2.3 in]{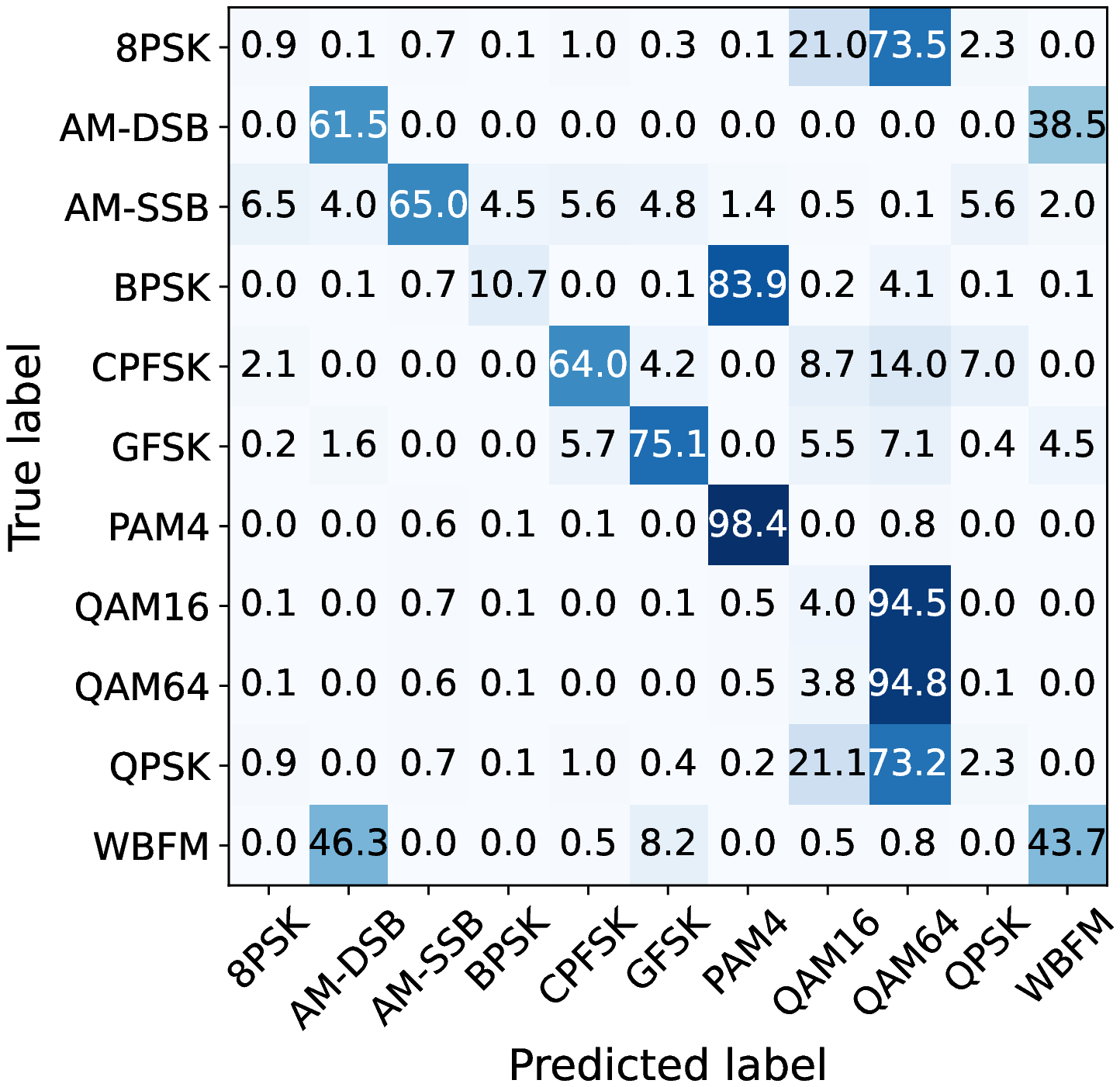}}
\caption{Overall confusion matrices, averaged over all five trials, from Experiment 1, train on low SNR data and test on high SNR data, showing the performance of the (left) CNN2, (middle) CNN2-257, and (right) Complex architectures. The respective average modulation pattern classification accuracies are 42.61\%, 41.24\%, and 47.31\%.}
\label{fig:conf1}
\end{figure*}

\begin{figure*}
 \centering
  \subfloat[\label{fig:conf2_me}]{%
       \includegraphics[width = 2.1 in]{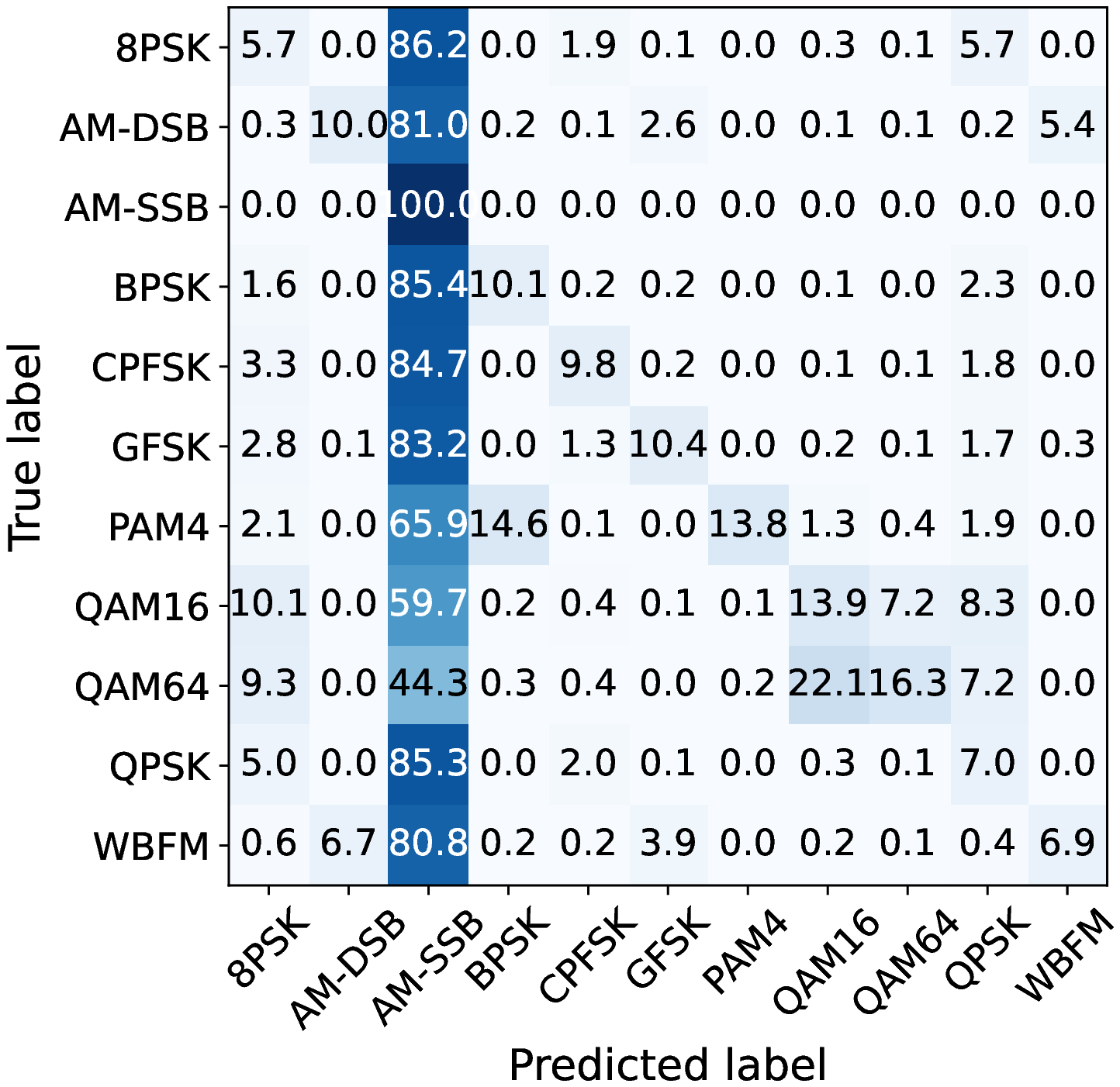}}
  \hfill
  \subfloat[\label{fig:conf2_1}]{%
        \includegraphics[width = 2.1 in]{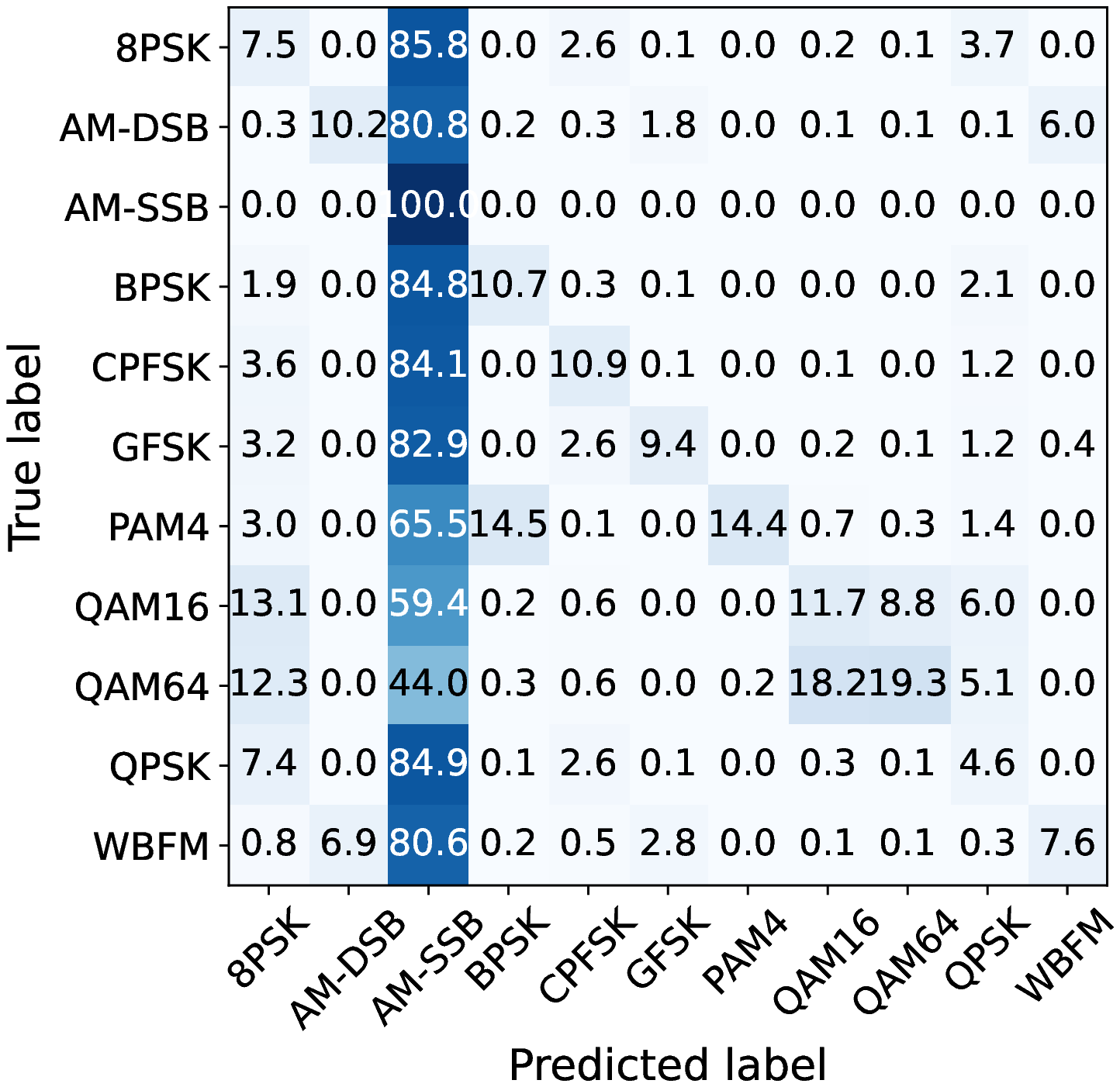}}
  \hfill  
  \subfloat[\label{fig:conf2_2}]{%
       \includegraphics[width = 2.1 in]{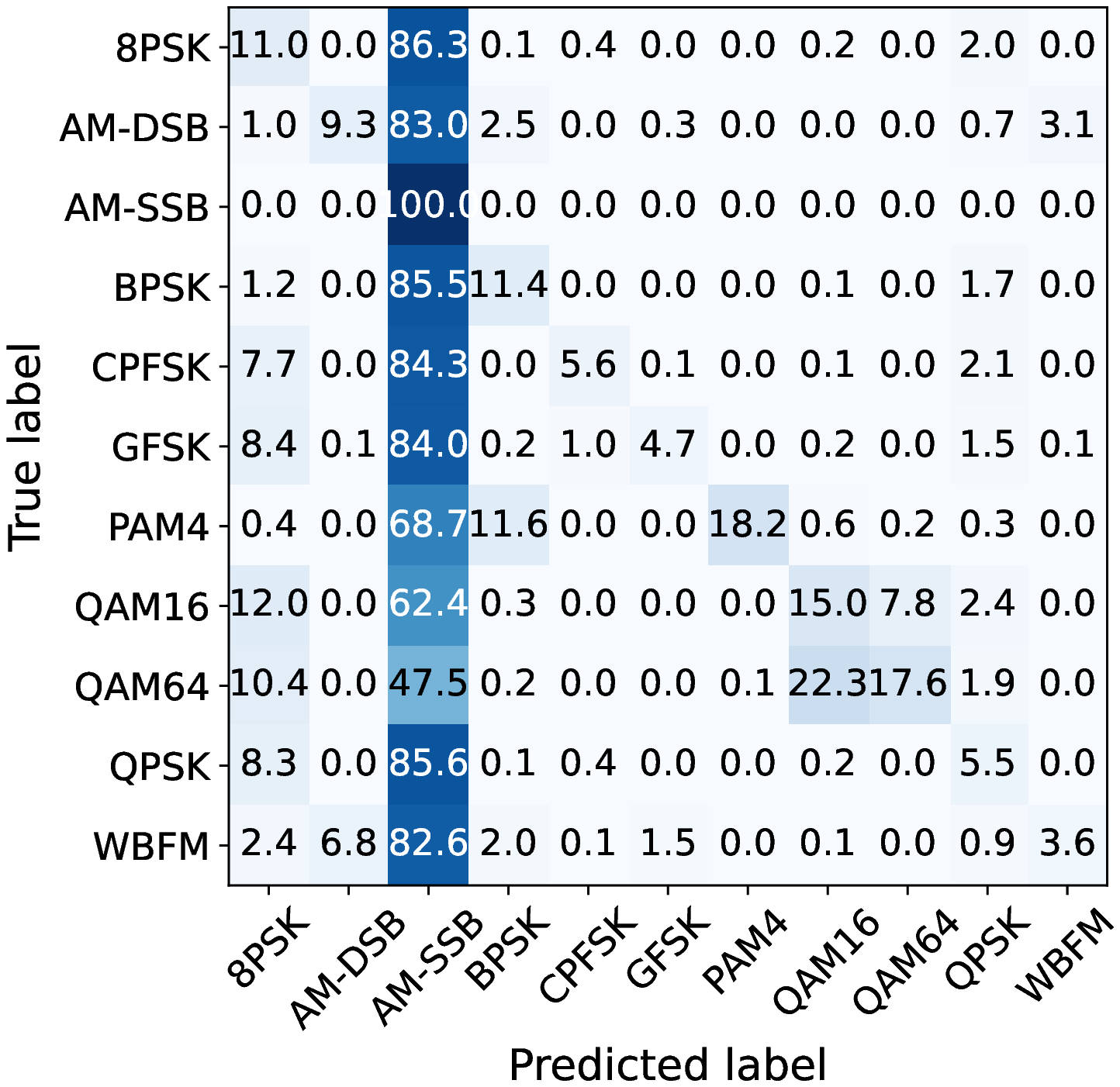}}
\caption{Overall confusion matrices from Experiment 2, train on high SNR data and test on low SNR data, showing the performance of the (left) CNN2, (middle) CNN2-257, and (right) Complex architectures. The respective modulation pattern classification accuracies are 18.53\%, 18.75\%, and 18.35\%.}
\label{fig:conf2}
\end{figure*}

Figures~\ref{fig:conf} (a-c),~\ref{fig:conf1} (a-c), and~\ref{fig:conf2} (a-c) display the averaged accuracy over all tested SNRs for each network in the three training paradigms respectively. In the train/test scheme of \cite{krzyston2020complex}, the CNN2 and CNN2-257 architectures had issues predicting QPSK modulation when it is an 8PSK modulation as well as predicting a WBFM modulation when it is the AM-DSB modulation, Figures~\ref{fig:conf}a and b. Figure~\ref{fig:conf}c shows the Complex architecture was able to better disambiguate these modulation patterns and more accurately classify them. However, all architectures showed a tendency to predict AM-SSB, regardless of the input.

In Experiment 1, seen in Figure~\ref{fig:conf1}, all the architectures show strong tendency to predict QAM64 for a variety of modulation patterns, namely QPSK, GFSK, CPFSK, and GFSK, Figure~\ref{fig:conf1}. However, the Complex network better disambiguated QAM64 samples, and performed almost three times as well for CPSFK samples and roughly twice as well for WBFM samples, Figure~\ref{fig:conf1}c. 

Figure~\ref{fig:conf2} highlights the difficulty all three architectures had in Experiment 2, when trained on high SNR data and tested on low SNR data. By design, CNNs are executing correlation processing by considering small portions of every input signal at an instance. Noise can overwhelm the ability of correlation processing to extract meaningful features from the noise. The ability to enhance a signal by correlation processing is the processing gain. With the addition of noise, the original signal level plus the processing gain is still much smaller than the noise, resulting in low classification accuracy.

\subsection{One-Hot Input Images}

\begin{figure*}
 \centering 
 \subfloat[\label{fig:max_me}]{%
 \includegraphics[width = 4 in]{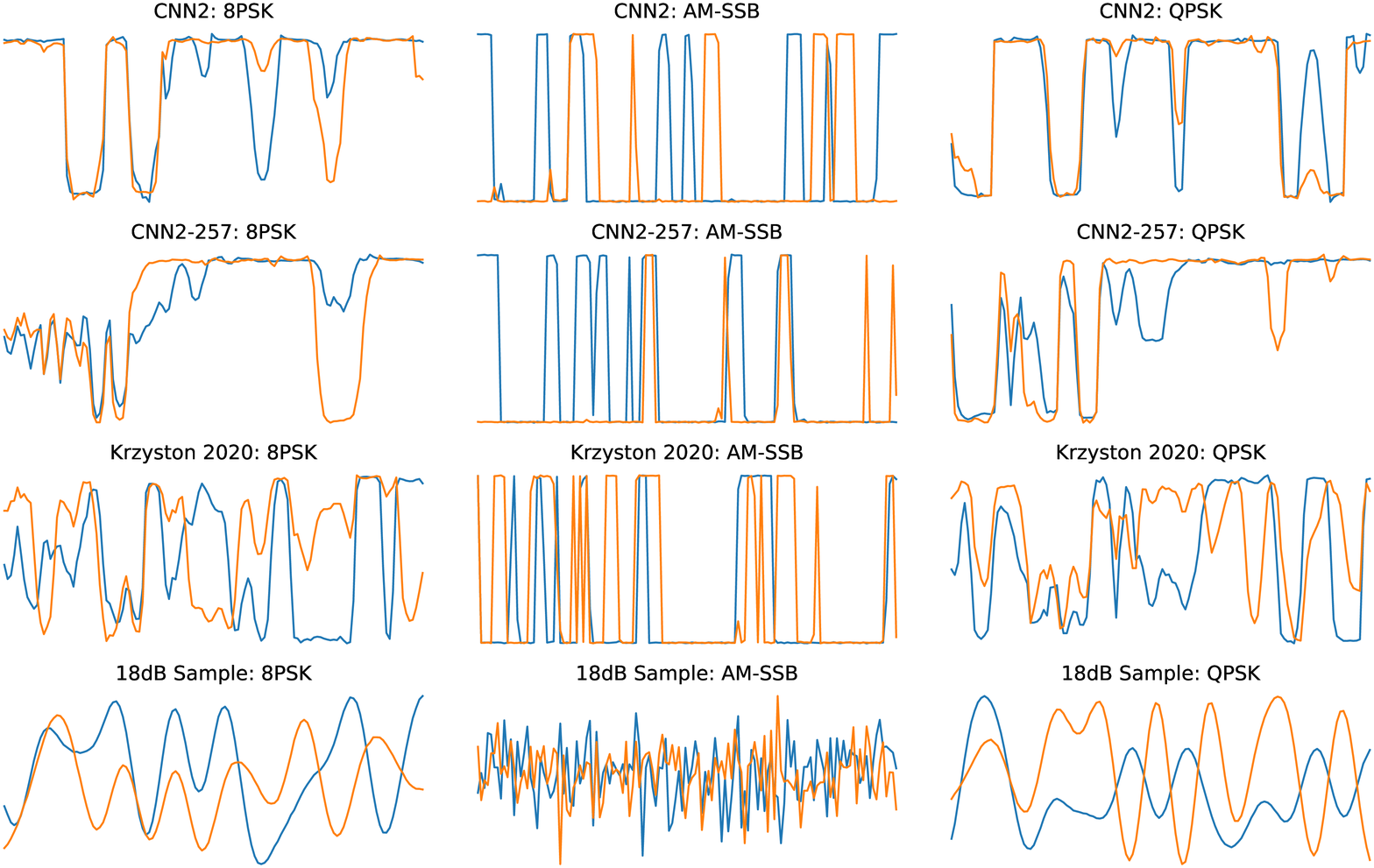}} \hfill \subfloat[\label{fig:max_1}]{%
 \includegraphics[width = 4 in]{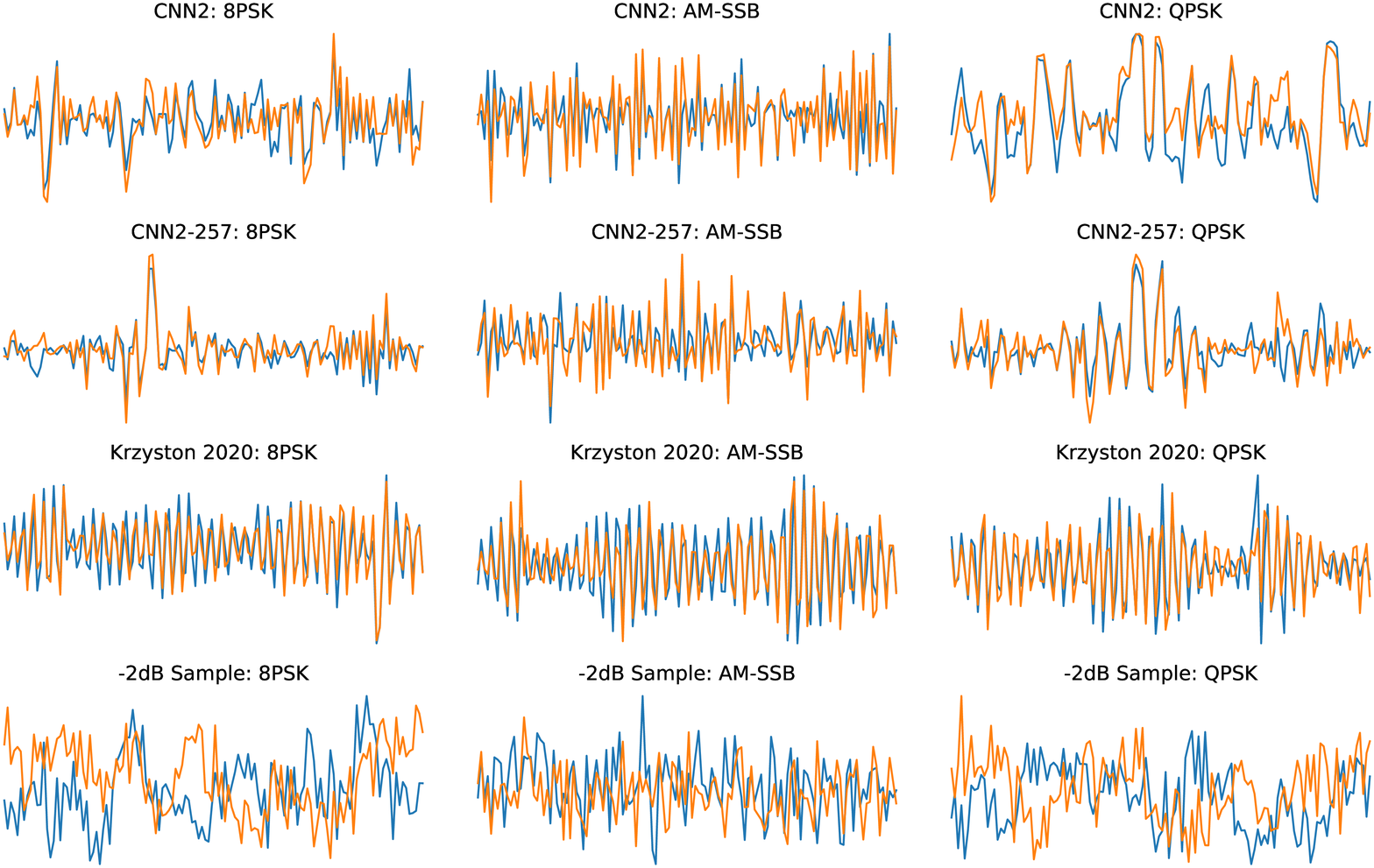}} \hfill \subfloat[\label{fig:max_2}]{%
 \includegraphics[width = 4 in]{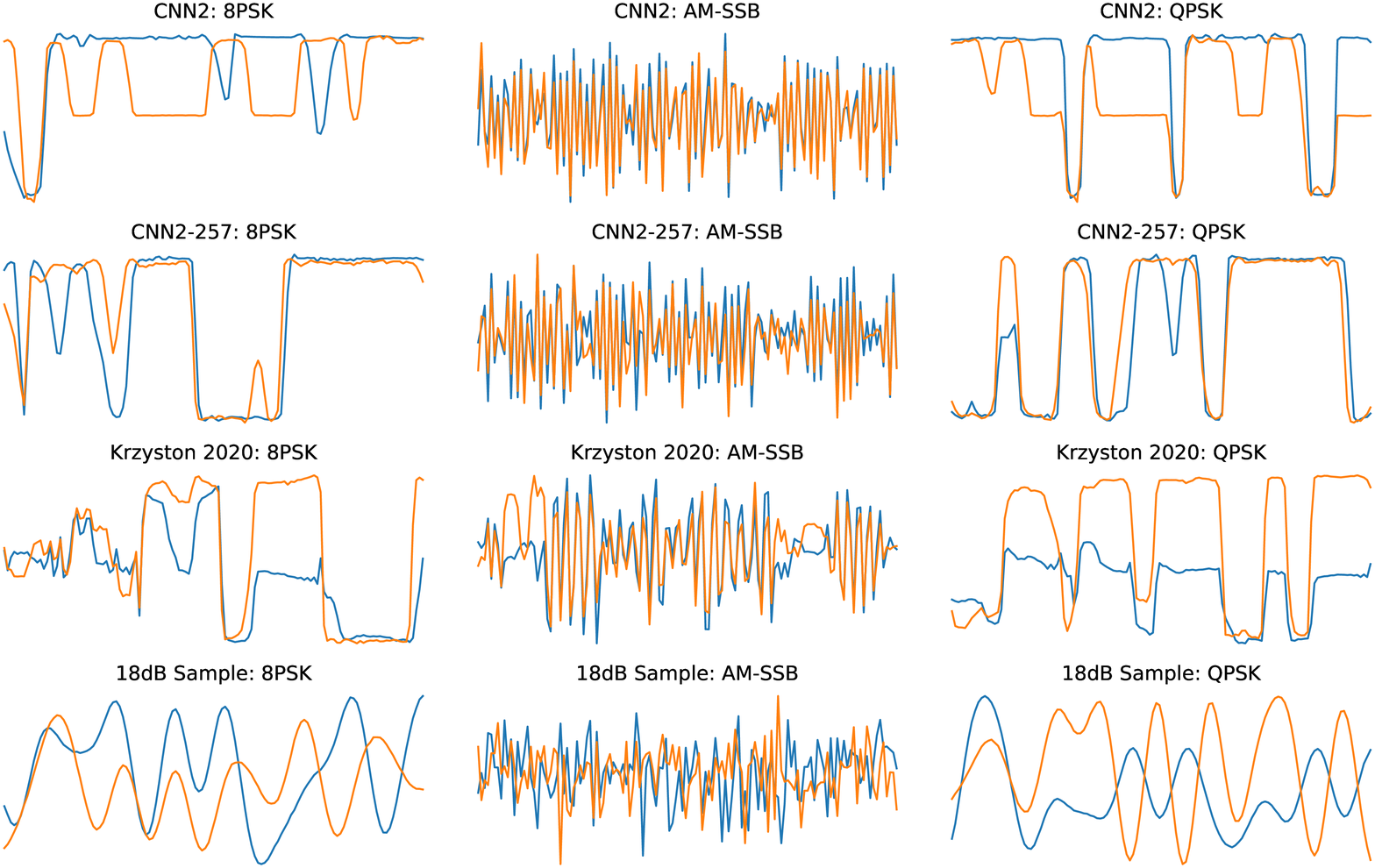}}\caption{Computed maximized input images for each architecture in (a) experiment performed in~\cite{krzyston2020complex}, (b) Experiment 1, and (c) Experiment 2. These images result in one-hot classification by the respective architecture, for the respective modulation pattern. Additionally, a sample from each modulation at the highest SNR included in the training set of that train/test paradigm is included for visual comparison.}
 \label{fig:max}
 \end{figure*}

A method used to help determine what a trained network has learned is the computation of the activation maximization for a specified neuron in a network \cite{erhan2009visualizing}. Activation maximization starts with a random input, and via backpropagation computes the input that maximizes the activation of the specified neuron. In CNNs, researchers have visualized filters from trained networks and computed the activation maximization to visualize the input those filters are detecting \cite{nguyen2015deep,simonyan2013deep}.

Here, for each of the three train/test paradigms, we computed the input that would result in one-hot modulation pattern classification for each architecture, shown in Figure~\ref{fig:max}. Each subplot contains an example of each modulation pattern, sampled from the highest SNR used in the training set of the train/test sequence. These figures demonstrate how well each network learns the structural features with reference to samples from the highest SNR part of the training sets.

In Figure~\ref{fig:max_me} are the one-hot activations for the experiment performed in \cite{krzyston2020complex}. In the BPSK and QPSK instances, the Complex network did produced an I/Q sample that showed similar relative magnitudes of the I and Q channels, and better captured both higher and lower frequency components of the modulation scheme. The AM-SSB activations for all networks were poor, in accordance with their tendency to incorrectly predict AM-SSB as shown in Figure~\ref{fig:conf}.

In Figure~\ref{fig:max_1}, are the one-hot activations for Experiment 1, train on low SNR data and test on high SNR data. It is apparent the high frequency noise in the training set had a significant impact on the quality of the representations learned by the architectures. Upon further investigation we notice the Complex networks ability to better capture the relative magnitudes of the I and Q channels and lower frequency content of the modulation pattern.

In Figure~\ref{fig:max_2}, are the one-hot activations for Experiment 2, train on high SNR data and test on low SNR data. The Complex network showed improved representation of the higher frequency content for the BPSK activation and QPSK activation.

We pose that enabling complex convolutions allows convolutional neural networks to learn time-domain features of the signal data as trajectories in a two-dimensional I/Q space. Treating the I and Q channels separately as pixels in an image neglects the interpretation as a trajectory in the 2D complex-plane. This conclusion is qualitatively derived from an examination of the one-hot images from the Complex network and comparing those to the same images from the CNN2 and CNN2-257 networks. Broadly speaking, the one-hot maximizing images more closely resemble the representative samples from the training data, meaning that the Complex network learned more of the time-domain structure of the data.

\section{Conclusion}We demonstrate the ability to compute complex convolutions in CNNs outperforms traditional CNNs in two train/test paradigms. Complex convolutions enables more robust features to be learned when trained and tested on all SNRs as well as when trained on low SNR data and tested on high SNR data, allowing for more robust pattern recognition. 

\section{Open Source Code}
The GitHub repository where code and information about the dataset can be found is github.com/JakobKrzyston/.

\bibliographystyle{IEEETran}
\bibliography{bib_JK}

\begin{thebibliography}{10}
\providecommand{\url}[1]{#1}
\csname url@samestyle\endcsname
\providecommand{\newblock}{\relax}
\providecommand{\bibinfo}[2]{#2}
\providecommand{\BIBentrySTDinterwordspacing}{\spaceskip=0pt\relax}
\providecommand{\BIBentryALTinterwordstretchfactor}{4}
\providecommand{\BIBentryALTinterwordspacing}{\spaceskip=\fontdimen2\font plus
\BIBentryALTinterwordstretchfactor\fontdimen3\font minus
  \fontdimen4\font\relax}
\providecommand{\BIBforeignlanguage}[2]{{%
\expandafter\ifx\csname l@#1\endcsname\relax
\typeout{** WARNING: IEEEtran.bst: No hyphenation pattern has been}%
\typeout{** loaded for the language `#1'. Using the pattern for}%
\typeout{** the default language instead.}%
\else
\language=\csname l@#1\endcsname
\fi
#2}}
\providecommand{\BIBdecl}{\relax}
\BIBdecl

\bibitem{isautier2015agnostic}
P.~Isautier, J.~Langston, J.~Pan, and S.~E. Ralph, ``Agnostic software-defined
  coherent optical receiver performing time-domain hybrid modulation format
  recognition,'' in \emph{Optical Fiber Communication Conference}.\hskip 1em
  plus 0.5em minus 0.4em\relax Optical Society of America, 2015, pp. Th2A--21.

\bibitem{isautier2015stokes}
P.~Isautier, J.~Pan, R.~DeSalvo, and S.~E. Ralph, ``Stokes space-based
  modulation format recognition for autonomous optical receivers,''
  \emph{Journal of Lightwave Technology}, vol.~33, no.~24, pp. 5157--5163,
  2015.

\bibitem{NIPS2012_4824}
\BIBentryALTinterwordspacing
A.~Krizhevsky, I.~Sutskever, and G.~E. Hinton, ``Imagenet classification with
  deep convolutional neural networks,'' in \emph{Advances in Neural Information
  Processing Systems 25}, F.~Pereira, C.~J.~C. Burges, L.~Bottou, and K.~Q.
  Weinberger, Eds.\hskip 1em plus 0.5em minus 0.4em\relax Curran Associates,
  Inc., 2012, pp. 1097--1105. [Online]. Available:
  \url{http://papers.nips.cc/paper/4824-imagenet-classification-with-deep-convolutional-neural-networks.pdf}
\BIBentrySTDinterwordspacing

\bibitem{huang2018gpipe}
Y.~Huang, Y.~Cheng, A.~Bapna, O.~Firat, M.~X. Chen, D.~Chen, H.~Lee, J.~Ngiam,
  Q.~V. Le, Y.~Wu, and Z.~Chen, ``Gpipe: Efficient training of giant neural
  networks using pipeline parallelism,'' 2018.

\bibitem{silver2017mastering}
D.~Silver, T.~Hubert, J.~Schrittwieser, I.~Antonoglou, M.~Lai, A.~Guez,
  M.~Lanctot, L.~Sifre, D.~Kumaran, T.~Graepel, T.~Lillicrap, K.~Simonyan, and
  D.~Hassabis, ``Mastering chess and shogi by self-play with a general
  reinforcement learning algorithm,'' 2017.

\bibitem{o2016convolutional}
T.~J. O’Shea, J.~Corgan, and T.~C. Clancy, ``Convolutional radio modulation
  recognition networks,'' in \emph{International conference on engineering
  applications of neural networks}.\hskip 1em plus 0.5em minus 0.4em\relax
  Springer, 2016, pp. 213--226.

\bibitem{krzyston2020complex}
J.~Krzyston, R.~Bhattacharjea, and A.~Stark, ``Complex-valued convolutions for
  modulation recognition using deep learning,'' in \emph{2020 IEEE
  International Conference on Communications Workshops (ICC Workshops)}.\hskip
  1em plus 0.5em minus 0.4em\relax IEEE, 2020, pp. 1--6.

\bibitem{o2018over}
T.~J. O’Shea, T.~Roy, and T.~C. Clancy, ``Over-the-air deep learning based
  radio signal classification,'' \emph{IEEE Journal of Selected Topics in
  Signal Processing}, vol.~12, no.~1, pp. 168--179, 2018.

\bibitem{ramjee2019fast}
S.~Ramjee, S.~Ju, D.~Yang, X.~Liu, A.~E. Gamal, and Y.~C. Eldar, ``Fast deep
  learning for automatic modulation classification,'' \emph{arXiv preprint
  arXiv:1901.05850}, 2019.

\bibitem{GoogLeNet}
\BIBentryALTinterwordspacing
C.~Szegedy, W.~Liu, Y.~Jia, P.~Sermanet, S.~Reed, D.~Anguelov, D.~Erhan,
  V.~Vanhoucke, and A.~Rabinovich, ``Going deeper with convolutions,'' in
  \emph{Computer Vision and Pattern Recognition (CVPR)}, 2015. [Online].
  Available: \url{http://arxiv.org/abs/1409.4842}
\BIBentrySTDinterwordspacing

\bibitem{arjovsky2016unitary}
M.~Arjovsky, A.~Shah, and Y.~Bengio, ``Unitary evolution recurrent neural
  networks,'' in \emph{International Conference on Machine Learning}, 2016, pp.
  1120--1128.

\bibitem{virtue2017better}
P.~Virtue, X.~Y. Stella, and M.~Lustig, ``Better than real: Complex-valued
  neural nets for mri fingerprinting,'' in \emph{2017 IEEE International
  Conference on Image Processing (ICIP)}.\hskip 1em plus 0.5em minus
  0.4em\relax IEEE, 2017, pp. 3953--3957.

\bibitem{trabelsi2017deep}
C.~Trabelsi, O.~Bilaniuk, Y.~Zhang, D.~Serdyuk, S.~Subramanian, J.~F. Santos,
  S.~Mehri, N.~Rostamzadeh, Y.~Bengio, and C.~J. Pal, ``Deep complex
  networks,'' \emph{arXiv preprint arXiv:1705.09792}, 2017.

\bibitem{chakraborty2019surreal}
R.~Chakraborty, Y.~Xing, and S.~Yu, ``Surreal: Complex-valued deep learning as
  principled transformations on a rotational lie group,'' \emph{arXiv preprint
  arXiv:1910.11334}, 2019.

\bibitem{erhan2009visualizing}
D.~Erhan, Y.~Bengio, A.~Courville, and P.~Vincent, ``Visualizing higher-layer
  features of a deep network,'' \emph{University of Montreal}, vol. 1341,
  no.~3, p.~1, 2009.

\bibitem{nguyen2015deep}
A.~Nguyen, J.~Yosinski, and J.~Clune, ``Deep neural networks are easily fooled:
  High confidence predictions for unrecognizable images,'' in \emph{Proceedings
  of the IEEE conference on computer vision and pattern recognition}, 2015, pp.
  427--436.

\bibitem{simonyan2013deep}
K.~Simonyan, A.~Vedaldi, and A.~Zisserman, ``Deep inside convolutional
  networks: Visualising image classification models and saliency maps,''
  \emph{arXiv preprint arXiv:1312.6034}, 2013.

\end{thebibliography}
\end{document}